\documentclass[10pt,twocolumn,letterpaper]{article}

\usepackage[pagenumbers]{cvpr} 
\usepackage[accsupp]{axessibility}

\makeatletter
\@namedef{ver@everyshi.sty}{}
\makeatother

\usepackage{graphicx}
\usepackage{amsmath}
\usepackage{amssymb}
\usepackage{booktabs}
\usepackage{tikz}
\usetikzlibrary{spy}
\usepackage{multirow}
\usepackage{subcaption}
\usepackage{overpic}

\newcommand{\myname}[0]{EVDI} %

\usepackage[pagebackref,breaklinks,colorlinks]{hyperref}

\usepackage[capitalize]{cleveref}
\crefname{section}{Sec.}{Secs.}
\Crefname{section}{Section}{Sections}
\Crefname{table}{Table}{Tables}
\crefname{table}{Tab.}{Tabs.}

\def\cvprPaperID{5205} 
\def\confName{CVPR}
\def\confYear{2022}

\begin{document}

\title{Unifying Motion Deblurring and Frame Interpolation with Events}

\author{Xiang Zhang,$\ $ Lei Yu\footnotemark[2]\\
Wuhan University, Wuhan, China.
\\
{\tt\small \{xiangz, ly.wd\}@whu.edu.cn}
}

\maketitle

\begin{abstract}
Slow shutter speed and long exposure time of frame-based cameras often cause visual blur and loss of inter-frame information, degenerating the overall quality of captured videos. To this end, we present a unified framework of event-based motion deblurring and frame interpolation for blurry video enhancement, where the extremely low latency of events is leveraged to alleviate motion blur and facilitate intermediate frame prediction. Specifically, the mapping relation between blurry frames and sharp latent images is first predicted by a learnable double integral network, and a fusion network is then proposed to refine the coarse results via utilizing the information from consecutive blurry inputs and the concurrent events. By exploring the mutual constraints among blurry frames, latent images, and event streams, we further propose a self-supervised learning framework to enable network training with real-world blurry videos and events. Extensive experiments demonstrate that our method compares favorably against the state-of-the-art approaches and achieves remarkable performance on both synthetic and real-world datasets. 
Codes are available at \url{https://github.com/XiangZ-0/EVDI}.

\end{abstract}

\section{Introduction}\label{Sec:Intro}
\footnotetext[2]{Corresponding author}
\footnotetext{The research was partially supported by the National Natural Science Foundation of China under Grants 61871297, the Natural Science Foundation of Hubei Province, China under Grant 2021CFB467, the Fundamental Research Funds for the Central University under Grant 2042020kf0019, and the National Natural Science Foundation of China Enterprise Innovation Development Key Project under Grant U19B2004.
}
Highly dynamic scenes, \eg, fast-moving targets or non-linear motions, pose challenges for high-quality video generation as the captured frame is often blurred and target information is missing between consecutive frames \cite{telleen2007synthetic}.
Existing frame-based methods attempt to tackle these problems by developing motion deblurring \cite{LEVS_jin2018learning}, frame interpolation \cite{dain_bao2019depth} or blurry video enhancement techniques \cite{flawless_jin2019learning,bin_shen2020blurry}. 
However, it is difficult for frame-based deblurring methods to predict sharp latent frames from severely blurred videos because of motion ambiguities and the erasure of intensity textures \cite{LEVS_jin2018learning}. 
Besides, current frame-based interpolation approaches generally assume the motion between neighboring frames to be linear \cite{dain_bao2019depth}, which is not always valid in real-world scenarios especially when encountering non-linear motions, thus often leading to incorrect predictions.

\begin{figure}[!t]
  \centering
  \includegraphics[width=0.95\linewidth]{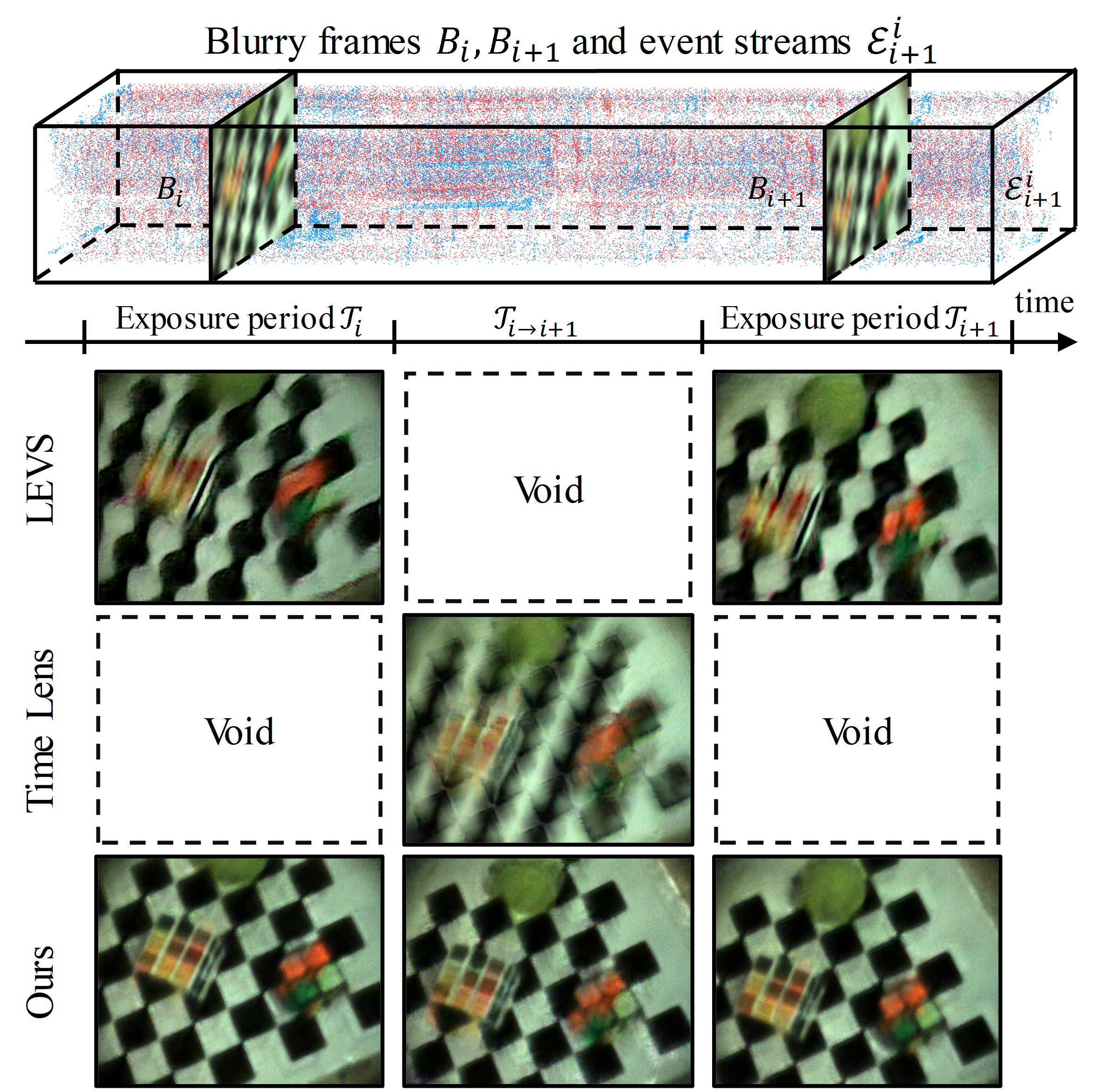}
    \vspace{-2mm}
  \caption{Illustrative examples of video deblurring and interpolation via the state-of-the-art deblurring approach LEVS \cite{LEVS_jin2018learning}, interpolation approach Time Lens \cite{timelens_tulyakov2021time} and our \myname\ method.
  }
  \label{Overview}
  \vspace{-5mm}
\end{figure}

\par
Recent works have revealed the advantages of event cameras \cite{survey_9138762} in motion deblurring and frame interpolation.
On one hand, the output of event camera inherently embeds precise motions and sharp edges \cite{benosman2013event} since it reports asynchronous event data with extremely low latency (in the order of $\mu s$) \cite{lichtsteiner128Times1282008,survey_9138762}, which is effective in alleviating motion blur \cite{edi_pan2019bringing,pan2020high,ledvdi_lin2020learning,esl_wang2020event,red_xu2021motion}. 
On the other hand, event camera is able to record almost continuous brightness changes to compensate the missing information between consecutive frames, making it feasible to recover accurate intermediate frames even under non-linear motions \cite{ledvdi_lin2020learning,timelens_tulyakov2021time}.
However, existing works generally treat motion deblurring and frame interpolation as separate tasks, while the problems of motion blur and missing information between frames have strong co-occurrence in real scenes and thus need to be considered simultaneously. In real-world scenarios, the aforementioned methods face two main challenges as follows.
\begin{itemize}
\setlength{\itemindent}{0em}
    \vspace{-1mm}
    \item \textbf{Limitations of Separate Tasks:} The performance of interpolation methods \cite{timelens_tulyakov2021time} is often highly dependent on the quality of reference frames, and it is difficult to interpolate clear results when the reference frames are degraded by motion blur. For deblurring task, most methods \cite{esl_wang2020event,red_xu2021motion} focus on recovering sharp images inside the exposure time of blurry inputs, neglecting these latent images between blurry frames (see Fig.~\ref{Overview}).
    \vspace{-5mm}
    \item \textbf{Data Inconsistency:} Most previous works employ well-labeled synthetic datasets for supervised \cite{esl_wang2020event,timelens_tulyakov2021time} or semi-supervised learning \cite{red_xu2021motion}, which often causes performance drop in real scenes due to the inconsistency between synthetic and real-world data \cite{red_xu2021motion}.
    \vspace{-1mm}
\end{itemize}

\par
In this paper, we present a unified framework of Event-based Video Deblurring and Interpolation (\myname) for blurry video enhancement. The proposed method consists of two main modules: a learnable double integral (LDI) network and a fusion network.
The LDI network is designed to automatically predict the mapping relation between blurry frames and sharp latent images from the corresponding events, where the timestamp of the latent image can be chosen arbitrarily inside the exposure time of blurry frames (deblurring task) or between consecutive blurry frames (interpolation task). 
The fusion network receives the coarse reconstruction of latent images and generates a fine result by utilizing all the information from consecutive blurry frames and event streams. 
For training, we take advantage of the mutual constraints among blurry frames, sharp latent images and event streams, and propose a fully self-supervised learning framework to help the network fit the distribution of real-world data without the need of ground truth images.


\par
The main contributions of this paper are three-fold:
\begin{itemize}
    \vspace{-2mm}
    \item We present a unified framework of event-based video deblurring and interpolation that generates arbitrarily high frame-rate sharp videos from blurry inputs.
    \vspace{-2mm}
    \item We propose a fully self-supervised framework to enable network training in real-world scenarios without any labeled data. 
    \vspace{-2mm}
    \item Experiments on both synthetic and real-world datasets show that our method achieves state-of-the-art results while maintaining an efficient network design.
\end{itemize}

\section{Related Work}\label{Sec:RelatedWork}
\subsection{Frame Interpolation}\label{subSec:Interpolation}
Existing frame-based interpolation methods can be roughly categorized into two classes: warping-based and kernel-based approaches. Warping-based approaches \cite{superslomo_jiang2018super,softmax_niklaus2020softmax,44_xue2019video,dain_bao2019depth} generally combine optical flow \cite{8flownet_ilg2017flownet,PWC_sun2018pwc} with image warping to predict intermediate frames, and several techniques have been proposed to enhance the interpolation performance, \eg, forward warping \cite{softmax_niklaus2020softmax}, spatial transformer networks \cite{44_xue2019video}, and depth information \cite{dain_bao2019depth}. However, these methods often assume linear motion and brightness constancy between two reference frames, thus failing to handle arbitrary motions.
Rather than warping reference frames with optical flow, kernel-based methods \cite{23AdaConv_niklaus2017video,24SepConv_niklaus2017video} model the frame interpolation as local convolution on the reference frames, where the kernel is directly estimated from the input frames. Despite kernel-based methods are more robust to complex motions and brightness changes, their scalability is often limited by the fixed sizes of convolution kernels.

\par
The common challenge of frame-based interpolation is the missing information between reference frames, which can be alleviated by leveraging the extremely low latency of events.
Recent approach \cite{timelens_tulyakov2021time} takes the merits of frames and events and achieves excellent interpolation results even under non-linear motions, but its performance is also closely related to the quality of reference frames and thus cannot be directly applied for blurry video enhancement.

\subsection{Motion Deblurring}\label{subSec:Deblurring}
One of the most popular frame-based deblurring methods is to employ neural networks to learn the blur feature and predict sharp images from blurry inputs \cite{LEVS_jin2018learning,8online_hyun2017online,42filter_zhou2019spatio,22intra_nah2019recurrent}. Several techniques have been developed to exploit the temporal information inside blurry frames, including dynamic temporal blending mechanism \cite{8online_hyun2017online}, spatio-temporal filter adaptive networks \cite{42filter_zhou2019spatio}, and intra-frame iterations \cite{22intra_nah2019recurrent}. 
Recent works have also revealed the potential of events in motion deblurring. Event streams inherently embed motion information and sharp edges, which can be exploited to tackle the temporal ambiguity and texture erasure caused by motion blur. Pioneer event-based methods achieve motion deblurring by relating blurry frames, sharp latent images and the corresponding events according to the physical event generation model \cite{edi_pan2019bringing,pan2020high}, but their performance is often degraded due to the imperfection of physical circuits, \eg, intrinsic camera noises. To alleviate this, learning-based approaches \cite{esl_wang2020event,red_xu2021motion} have been proposed to fit the distribution of event data, achieving better deblurring performance.
\par
However, most deblurring methods focus solely on restoring sharp latent images inside the exposure time of blurry frames, while the information between blurry frames is also important in practical applications, motivating the combination of deblurring and interpolation.

\subsection{Joint Deblurring and Interpolation}\label{subSec:Joint}
Previous frame-based methods have approached the joint deblurring and interpolation task \cite{flawless_jin2019learning,bin_shen2020blurry}. The work of \cite{flawless_jin2019learning} performs frame interpolation based on the keyframes pre-processed by a deblurring module, and the work of \cite{bin_shen2020blurry} treats deblurring and interpolation as a unified task and achieves better enhancement performance. For event-based methods, LEDVDI is the closest related work \cite{ledvdi_lin2020learning}, but LEDVDI is categorized into the cascaded scheme as it achieves deblurring and interpolation with different stages. Besides, all the aforementioned methods require supervised training on the synthetic datasets, limiting their performance on real-world scenarios due to data inconsistency. 

\par 
Exploiting the information from both frames and events, our method achieves blurry video enhancement without distinguishing the deblurring and interpolation tasks. 
Moreover, a self-supervised learning framework is proposed to enable network training with real events and blurry videos, guaranteeing the performance in real-world scenarios.


\section{Problem Statement}\label{Sec:ProbState}
Videoing highly dynamic scenes often suffers from blurry artifacts and the Blurry Video Enhancement (BVE) plays an important role for visual perception. 
Existing frame-based methods often struggle to achieve BVE due to motion ambiguity and loss of inter-frame information, while this can be effectively mitigated with the aid of events.
Given two consecutive blurry frames $B_i, B_{i+1}$ captured within the exposure time $\mathcal{T}_i, \mathcal{T}_{i+1}$ and the corresponding event streams $\mathcal{E}_{i+1}^{i}$ triggered inside $\mathcal{T}_{i+1}^{i}$, where $\mathcal{T}_{i+1}^{i}\triangleq \mathcal{T}_i\cup \mathcal{T}_{i\rightarrow i+1}\cup \mathcal{T}_{i+1}$ with $\mathcal{T}_{i\rightarrow i+1}$ indicating the time interval between $B_i$ and $B_{i+1}$, the task of \myname\ is to achieve BVE directly from blurry inputs, \ie,
\begin{equation}\label{eq:evdi}
    L(t) = \operatorname{\myname}(t; B_i, B_{i+1},\mathcal{E}_{i+1}^{i}),\quad \forall t\in \mathcal{T}_{i+1}^{i},
\end{equation}
where $L(t)$ indicates the latent image of arbitrary time $t \in \mathcal{T}_{i+1}^{i}$. According to \cref{eq:evdi}, \myname\ degrades to Motion Deblurring (MD) when $t\in \mathcal{T}_i$ or $\mathcal{T}_{i+1}$, or Frame Interpolation (FI) when $t\in \mathcal{T}_{i\to i+1}$. Thus, \myname\ is more general than MD and FI, and provides a unified formulation to the task of BVE.

\par 
\noindent \textbf{{\myname} \vs Frame Interpolation.} Conventional FI task aims at recovering the intermediate latent images $\{L(t)\}_{t\in \mathcal{T}_{i \rightarrow i+1}}$ from sharp reference frames $I_i,\ I_{i+1}$. Providing the concurrent event streams $\mathcal{E}_{i \rightarrow i+1}$ emitted within $\mathcal{T}_{i\rightarrow i+1}$, we have
\begin{equation}\label{Interp}
    L(t) = \operatorname{Interp}(t; I_i, I_{i+1},\mathcal{E}_{i \rightarrow i+1}),\quad t\in \mathcal{T}_{i \rightarrow i+1},
\end{equation}
where $\operatorname{Interp}(\cdot)$ represents an FI operator. Most FI methods \cite{superslomo_jiang2018super,dain_bao2019depth,timelens_tulyakov2021time} are designed to restore inter-frame latent images from high-quality (sharp and clear) reference frames $I_i,\ I_{i+1}$, while \myname\ directly accepts blurry inputs, which is more challenging than conventional FI.

\par 
\noindent \textbf{{\myname} \vs Motion Deblurring.} The MD aims at reconstructing the sharp latent images $\{L(t)\}_{t\in \mathcal{T}_i}$ from the corresponding blurry frame $B_i$. Providing the concurrent event streams $\mathcal{E}_i$ triggered within $\mathcal{T}_i$, we have
\begin{equation}\label{Deblur}
    L(t) = \operatorname{Deblur}(t; B_i, \mathcal{E}_i),\quad t\in \mathcal{T}_i,
\end{equation}
where $\operatorname{Deblur}(\cdot)$ indicates an MD operator. Existing deblurring methods \cite{Deng_2021_ICCV,Suin_2021_CVPR} mainly focus on recovering the latent frames inside the exposure time $\mathcal{T}_i$, while \myname\ is able to predict the latent images of time instance both inside the exposure time $\mathcal{T}_i$ (or $\mathcal{T}_{i+1}$) and between blurry frames $\mathcal{T}_{i \rightarrow i+1}$, as shown in Fig.~\ref{Overview}. 

\par
Ideally, \myname\ can approach the BVE task by unifying MD and FI in \cref{eq:evdi}. However, to efficiently realize \myname\ in real-world scenarios, challenges still exist.

\begin{itemize}
    \vspace{-1mm}
    \item MD and FI should be simultaneously addressed in a unified framework to fulfill the \myname. Previous attempts for BVE \cite{flawless_jin2019learning, ledvdi_lin2020learning} employ a cascaded scheme that performs frame interpolation after deblurring, but this approach often propagates deblurring error to the interpolation stage, leading to sub-optimal results.
    \vspace{-1mm}
    \item Existing related methods are generally developed within a supervised learning framework  \cite{ledvdi_lin2020learning,flawless_jin2019learning,bin_shen2020blurry}, of which the supervision is usually provided by synthetic blurry images and events. Thus the performance might degrade in real scenes due to the different distribution between synthetic and real-world data.
\end{itemize}

\section{Method}
In this work, we propose to approximate the optimal \myname\ model with trainable neural networks, and develop a self-supervised learning framework by exploiting the mutual constraints among blurry frames, sharp latent frames and event streams.

\subsection{Unified Deblurring and Interpolation}
We first review the physical generation model of events, which are triggered whenever the log-scale brightness change exceeds the event threshold $c>0$, \ie,
\begin{equation}
    \operatorname{log}(L(t,\mathbf{x})) - \operatorname{log}(L(\tau,\mathbf{x})) = p\cdot c,
\end{equation}
where $L(t,\mathbf{x})$ and $L(\tau,\mathbf{x})$ denote the instantaneous intensity at time $t$ and $\tau$ at the pixel position $\mathbf{x}$, and polarity $p\in\{+1,-1\}$ indicates the direction of brightness changes. With the aid of events, we can formulate the following relation (the pixel position is omitted for readability):
\begin{equation}\label{latent_expEvent}
    L(t) = L(f) \operatorname{exp}(c\int_f^t e(s)ds),
\end{equation}
where $L(t)$ and $L(f)$ are latent images at instant time $t$ and $f$, and $e(t)\triangleq p \cdot \delta(t-\tau)$ denotes the continuous representation of events with $\delta(\cdot)$ indicating the Dirac function. On the other hand, the blurry images can be formulated as the average of latent images within the exposure time \cite{chen2018reblur2deblur}, \ie,
\begin{equation}\label{blurry_latent}
    B = \frac{1}{T} \int_{t\in \mathcal{T}} L(t) dt
\end{equation}
with $T$ denoting the duration of exposure period $\mathcal{T}$.
Combining Eq.~\eqref{latent_expEvent} and Eq.~\eqref{blurry_latent}, one can obtain
\begin{align}
     L(f) &= \frac{B}{E(f,\mathcal{T})} , \quad \text{with} \label{blurry_latent_event} 
    \\
    E(f,\mathcal{T}) &= \frac{1}{T} \int_{t \in \mathcal{T}} \operatorname{exp} (c\int_f^t e(s)ds)dt \label{EDI} 
\end{align}
representing the relation between blurry frames $B$ and latent images $L(f)$ from the perspective of events, which is also known as event-based double integral (EDI) \cite{edi_pan2019bringing}.

\subsubsection{Feasibility Analysis}
Previous works of  \cite{edi_pan2019bringing,pan2020high} focus on exploiting Eq.~\eqref{blurry_latent_event} to restore the sharp latent image inside the exposure period $\mathcal{T}$, while this formulation can be also extended to recover the latent frames at arbitrary time outside $\mathcal{T}$ (please see the supplementary material for proof). However, direct applying Eq.~\eqref{blurry_latent_event} for unified deblurring and interpolation often meets the following obstacles:
First, the computation of $E(f,\mathcal{T})$ requires the knowledge of event threshold $c$, which is critical to the recovery performance \cite{edi_pan2019bringing} but hard to accurately estimate due to its temporal instability. Second, real-world events are noisy due to the non-ideality of physical sensors \cite{survey_9138762}, \eg, limited read-out bandwidth, and thus often lead to degraded results, especially when encountering long-term integral of events where $E(f,\mathcal{T})$ is severely contaminated by noises. Hence, we propose to employ learning-based approaches to fit the statistics of real-world events.

\begin{figure}[!t]
  \centering
    \includegraphics[width=1\linewidth]{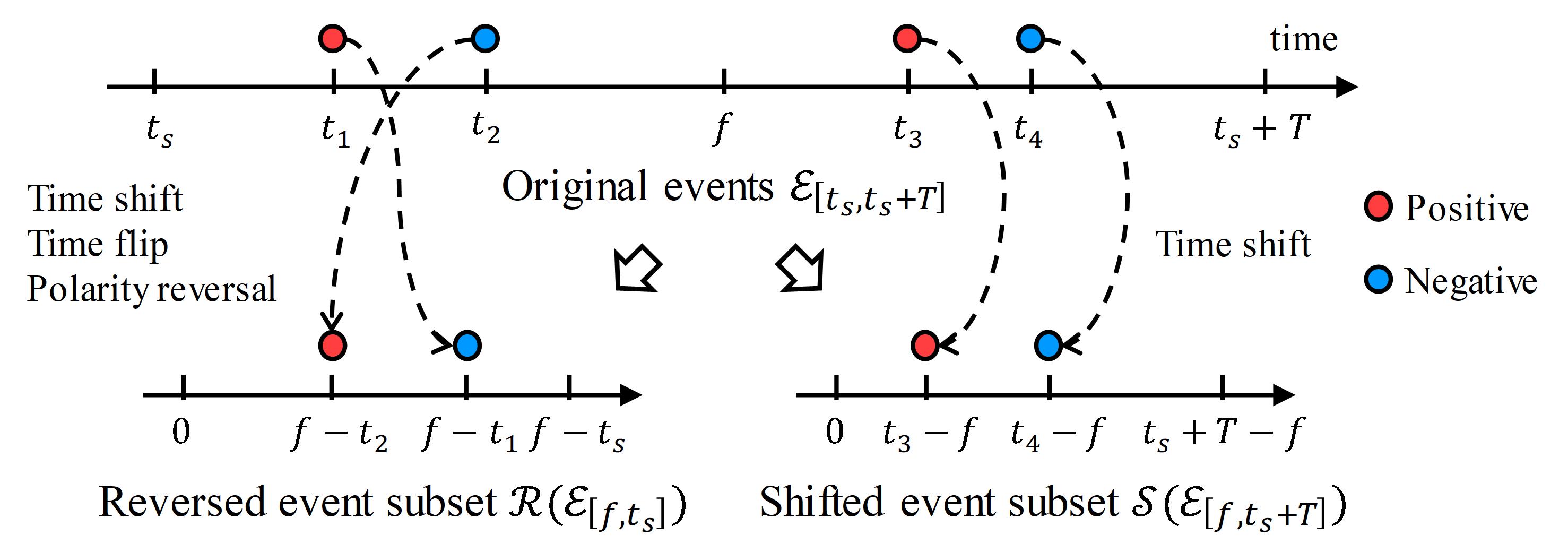}
    \vspace{-2mm}
    \caption{Example of the pre-processing operation. The left event subset experiences time shift, flip and polarity reversal, \ie, $\mathcal{R}(\cdot)$,  since $t_s-f<0$, and the right subset is processed by the operator of time shift, \ie, $\mathcal{S}(\cdot)$, as $t_s+T-f \geq 0$.}
    \label{PreExample}
    \vspace{-3mm}
\end{figure}
\subsubsection{Network Architecture}
\begin{figure}[!t]
  \centering
   \includegraphics[width=1\linewidth]{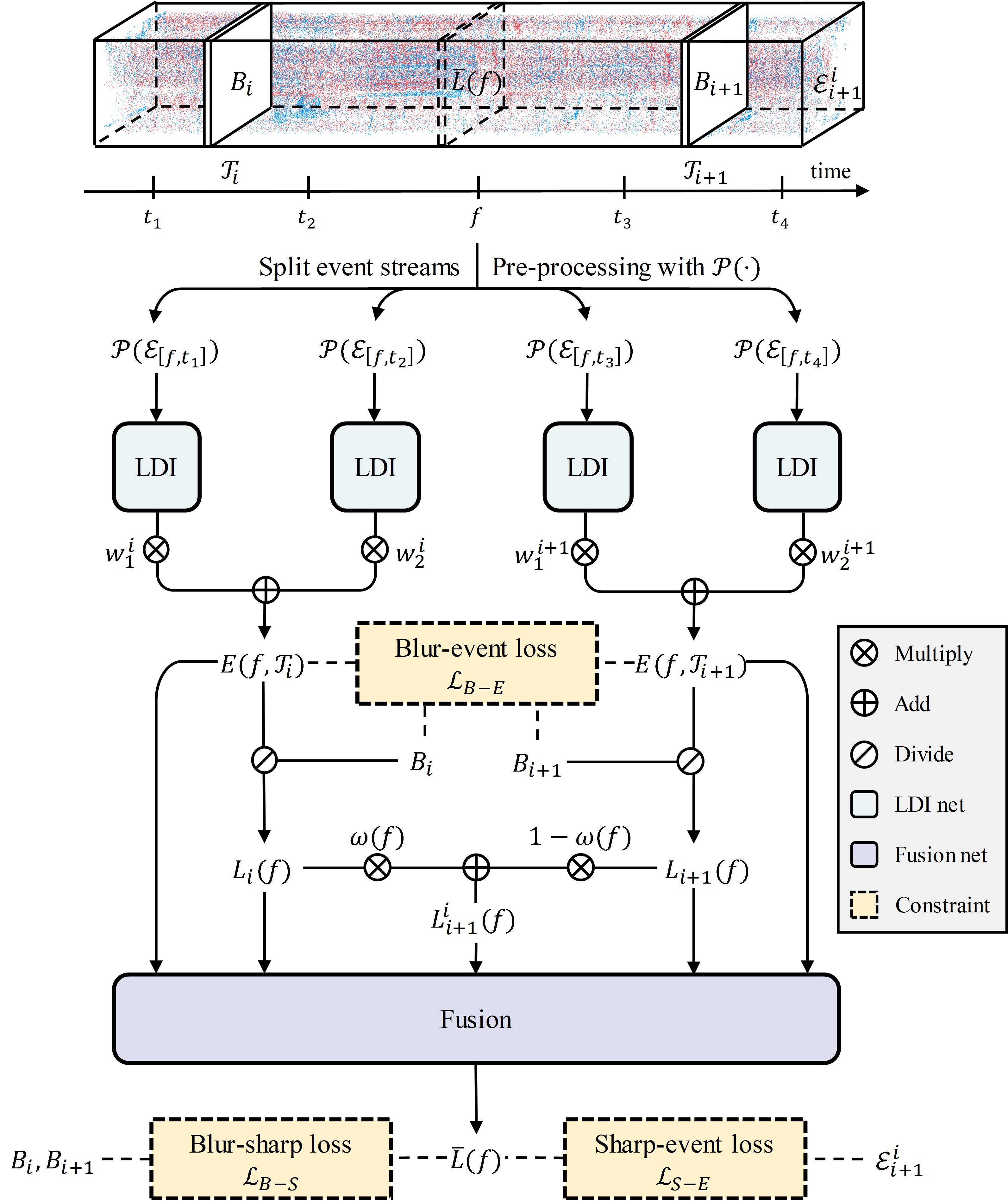}
    \vspace{-2mm}
   \caption{Data flow of the proposed method. With two consecutive blurry frames $B_i,B_{i+1}$ and the concurrent events $\mathcal{E}_{i+1}^{i}$, we first split the events into four subsets according to the target timestamp $f$, and then feed the events pre-processed by $\mathcal{P}(\cdot)$ to four weight-sharing LDI networks. Following that, three coarse results $L_i(f), L_{i+1}(f), L_{i+1}^{i}(f)$ are generated based on $B_i,B_{i+1}$ and the corresponding double integral of events $E(f,\mathcal{T}_i), E(f,\mathcal{T}_{i+1})$, and the final result $\bar{L}(f)$ is produced by refining them with the fusion network.}
   \label{Network}
   \vspace{-3mm}
\end{figure}

Our network receives a latent image timestamp $f\in\mathcal{T}_{i+1}^{i}$, two consecutive blurry frames $B_i, B_{i+1}$ and the corresponding event streams $\mathcal{E}_{i+1}^{i}$ as input, and outputs a sharp latent image $L(f)$. There are two main modules in our network: a learnable double integral (LDI) network and a fusion network, where the LDI network learns to approximate the double integral behavior of Eq.~\eqref{EDI} and the fusion network is designed to refine the results generated by the blurry images and the outputs of LDI network, as shown in Fig.~\ref{Network}.

\par 
\noindent \textbf{LDI Network.}
Suppose the LDI network is trained to approximate a specific case $E(0,\mathcal{T}_{[0,T]}) \approx \operatorname{LDI}(\mathcal{E}_{[0,T]})$ \ie,
\begin{equation}\label{LDI}
    \operatorname{LDI}(\mathcal{E}_{[0,T]}) \approx \frac{1}{T} \int_0^T \operatorname{exp}(c \int_0^t e(s) ds) dt,
\end{equation}
where $\mathcal{T}_{[0,T]}$ indicates the time interval from $0$ to $T>0$ and $\mathcal{E}_{[0,T]}$ is the corresponding event streams.
Now we consider a more general case of $E(f,\mathcal{T})$, which can be written as
\begin{equation}\label{EDI_divide}
\begin{aligned}
     E(f,\mathcal{T}) =& \frac{1}{T}  \int_{t_s}^{f} \operatorname{exp} (c\int_f^t e(s)ds)dt 
     \\
     &+ \frac{1}{T} \int_{f}^{t_s+T} \operatorname{exp} (c\int_f^t e(s)ds)dt,
\end{aligned}
\end{equation}
where $t_s$ indicates the starting time of $\mathcal{T}$. Applying $t'=t-f$ and $s'=s-f$ to Eq.~\eqref{EDI_divide}, we have 
\begin{equation}\label{E_2G}
\begin{aligned}
    E(f,\mathcal{T}) =& -\frac{1}{T} \int_0^{t_s - f} \operatorname{exp}(c\int_0^{t'} e(s'+f)ds')dt'
    \\
    &+ \frac{1}{T} \int_0^{t_s+T-f} \operatorname{exp}(c\int_0^{t'} e(s'+f)ds')dt'
    \\
    =& w_1 G(\mathcal{E}_{[f,t_s]}) + w_2 G(\mathcal{E}_{[f,t_s+T]}),
\end{aligned}
\end{equation}
where $w_1=(f-t_s)/T,\ w_2=(t_s+T-f)/T$ are weights and $G(\cdot)$ is a general formula defined as
\begin{equation}\label{G}
    G(\mathcal{E}_{[f,t_r]}) = \frac{1}{t_r-f}\int_0^{t_r-f} \operatorname{exp}(c\int_0^t e(s+f) ds) dt
\end{equation}
and $t_r$ denotes the reference time. Based on the above definition, we can calculate $E(f,\mathcal{T})$ by approximating Eq.~\eqref{G} with the LDI network, \ie, Eq.~\eqref{LDI}. For the case of $t_r-f\geq 0$, $G(\cdot)$ can be directly approximated by 
\begin{equation}
    G(\mathcal{E}_{[f,t_r]}) \approx \operatorname{LDI}(\mathcal{S}(\mathcal{E}_{[f,t_r]}))
\end{equation}
with $\mathcal{S}(\mathcal{E}_{[f,t_r]}) \triangleq \{e(t+f),t\in[0,t_r-f]\}$ representing the event operator of time shift. For the case of $t_r-f< 0$, 
\begin{equation}
\begin{aligned}
    G(\mathcal{E}_{[f,t_r]}) &= \frac{1}{f-t_r} \int_0^{f-t_r} \operatorname{exp}(c\int_0^t -e(-s+f) ds)dt
    \\ 
    &\approx \operatorname{LDI}(\mathcal{R}(\mathcal{E}_{[f,t_r]})),
\end{aligned}
\end{equation}
where $\mathcal{R}(\mathcal{E}_{[f,t_r]}) \triangleq \{-e(-t+f),t\in[0,f-t_r]\}$ indicates the event operator composed of time shift, flip and polarity reversal, as shown in Fig.~\ref{PreExample}. For simplicity, we define a unified pre-processing operator $\mathcal{P}(\cdot)$ as follows.
\begin{equation}\label{P}
    \mathcal{P}(\mathcal{E}_{[f,t_r]}) =\left\{
	\begin{array}{ll}
	\mathcal{S}(\mathcal{E}_{[f,t_r]}) & \text { if } t_r \geq f, \\
	\mathcal{R}(\mathcal{E}_{[f,t_r]}) & \text { if } t_r < f.
	\end{array}\right.
\end{equation}
Thus, Eq.~\eqref{E_2G} can be reformulated as
\begin{equation}
    E(f,\mathcal{T}) \approx w_1 \operatorname{LDI}(\mathcal{P}(\mathcal{E}_{[f,t_s]})) + w_2 \operatorname{LDI}(\mathcal{P}(\mathcal{E}_{[f,t_s+T]})),
\end{equation}
meaning that arbitrary $E(f,\mathcal{T})$ can be approximated by a weighted combination of LDI outputs, where the LDI network only needs to be trained once to fit the case of Eq.~\eqref{LDI}.

\par 
For the input of LDI network, we introduce a spatio-temporal event representation.
With a pre-defined number, \eg, $N$, we fairly divide $N$ temporal bins from $t=0$ to $t=T_{i+1}^{i}$ where $T_{i+1}^{i}$ denotes the total duration of $\mathcal{T}_{i+1}^{i}$. We then accumulate the events pre-processed by $\mathcal{P}(\cdot)$ inside each temporal bin, and form a $2N \times H \times W$ tensor as the LDI input with $2,H,W$ indicating event polarity, image height and width, respectively. Therefore, our event representation enables flexible choice of the target timestamp $f$ while maintaining a fixed input format, which allows network to restore the latent images $L(f)$ at arbitrary $f\in \mathcal{T}_{i+1}^{i}$ without any network modification or re-training process.

\par 
\noindent \textbf{Fusion Network.} After obtaining $E(f,\mathcal{T})$ from LDI network, the latent image $L(f)$ can be coarsely restored by Eq.~\eqref{blurry_latent_event}.
We denote the latent images reconstructed from $B_i, B_{i+1}$ as $L_i(f), L_{i+1}(f)$, respectively, and manually generate an extra result $L^{i}_{i+1}(f)$ by
\begin{equation}
    L^{i}_{i+1}(f) = \omega(f) L_i(f) + (1-\omega(f)) L_{i+1}(f),
\end{equation}
where the weighting function $\omega(f)$ with $f \in [0,T_{i+1}^i]$ is defined as 
\begin{equation}
    \omega(f) =\left\{
	\begin{array}{ll}
	1 & \text { if } f \in \mathcal{T}_i, 
	\\
	1 - \frac{f}{T_{i+1}^i} & \text { if } f \in \mathcal{T}_{i\rightarrow i+1},
	\\
	0 & \text { if } f \in \mathcal{T}_{i+1},
	\end{array}\right.
\end{equation}
since a weighted reconstruction is helpful for frame interpolation in our observation. Finally, our fusion network receives $L_i(f), L_{i+1}(f), L^{i}_{i+1}(f), E(f,\mathcal{T}_i), E(f,\mathcal{T}_{i+1})$ and produces the final latent image $\bar{L}(f)$, as illustrated in Fig.~\ref{Network}.


\input{Figs-Fig-deblur}
\begin{table*}[th]
\centering
\small
\caption{Quantitative comparisons of the proposed method to the state-of-the-arts on the deblurring task. Note that LEDVDI only produces 6 frames for each blurry frame while the others output 7 frames. The number of network parameters (\#Param.) is also provided.}
\vspace{-3mm}
\begin{tabular}{lcccccccc}
\hline
\multirow{2}{*}{Method} & \multicolumn{3}{c}{GoPro} &  & \multicolumn{3}{c}{HQF}                           & \multirow{2}{*}{\#Param.} \\ \cline{2-4} \cline{6-8}
                        & PSNR $\uparrow$   & SSIM $\uparrow$   & LPIPS $\downarrow$  &  & PSNR $\uparrow$          & SSIM $\uparrow$           & LPIPS $\downarrow$           &                           \\ \hline
LEVS \cite{LEVS_jin2018learning}                    & 20.84  & 0.5473  & 0.1111 &  & 20.08          & 0.5629          & 0.0998          & 18.21M                    \\
EDI \cite{edi_pan2019bringing}                     & 21.29  & 0.6402  & 0.1104 &  & 19.65          & 0.5909          & 0.1173          & -                         \\
LEDVDI \cite{ledvdi_lin2020learning}                  & \underline{25.38}  & 0.8567  & \underline{0.0280} &  & 22.58          & 0.7472          & 0.0578          & 4.996M                    \\
eSL-Net \cite{esl_wang2020event}                 & 17.80  & 0.5655  & 0.1141 &  & 21.36          & 0.6659          & 0.0644          & \textbf{0.188M}           \\
RED \cite{red_xu2021motion}                     & 25.14  & \underline{0.8587}  & 0.0425 &  & \underline{24.48}          & \underline{0.7572}          & \underline{0.0475}          & 9.762M                    \\ \hline
\myname\ (Ours)                    & \textbf{30.40}       &  \textbf{0.9058}       & \textbf{0.0144}       &  & \textbf{24.77} & \textbf{0.7664} & \textbf{0.0423} & \underline{0.393M}                    \\ \hline
\end{tabular}
\label{tab:deblur}
\vspace{-3mm}
\end{table*}
\subsection{Self-supervised Learning Framework}
The proposed self-supervised learning framework consists of three different losses which are formulated based on the mutual constraints among blurry frames, sharp latent images and event streams.

\par 
\noindent \textbf{Blurry-event Loss.} 
The double integral of events $E(f,\mathcal{T})$ corresponds to the mapping relation between blurry frames and sharp latent images. For multiple blurry inputs, we propose to formulate the consistency between the latent images reconstructed from different blurry frames, \eg, $L_i(f) = L_{i+1}(f)$. Considering the quantization error which might be accumulated in $E(f,\mathcal{T})$, we rewrite the consistency as
\begin{equation}\label{E2B2-div}
    \frac{B_i}{E(f,\mathcal{T}_i)} \approx \frac{B_{i+1}}{E(f,\mathcal{T}_{i+1})},
\end{equation}
where $E(f,\mathcal{T}_i), E(f,\mathcal{T}_{i+1})$ are generated by the LDI network. We convert Eq.~\eqref{E2B2-div} to the logarithmic domain and rewrite it as the blurry-event loss $\mathcal{L}_{B\text{-}E}$,
\begin{equation}
    \mathcal{L}_{B\text{-}E} = \|(\tilde{B}_{i+1} - \tilde{B}_{i}) - (\tilde{E}(f,\mathcal{T}_{i+1}) - \tilde{E}(f,\mathcal{T}_{i})) \|_1,
\end{equation}
where the top tilde denotes logarithm, \eg, $\tilde{B}_{i} = \operatorname{log}(B_i)$. With the blurry-event constraint, the LDI network can learn to perform event double integral through utilizing the brightness difference between blurry frames. 

\par 
\noindent \textbf{Blurry-sharp Loss.} 
Providing the reconstructed latent images $\bar{L}(t)$ with $t \in \mathcal{T}_i$, the blurring process Eq.~\eqref{blurry_latent} can be reformulated as the discrete version, \ie,
\begin{equation}\label{reblur}
    \bar{B}_i = \frac{1}{T} \int_{t\in \mathcal{T}_i} \bar{L}(t) dt \approx \frac{1}{M} \sum_{m=0}^{M-1} \bar{L}_{i} [m],
\end{equation}
where $\bar{L}_{i} [m]$ indicates the $m$-th latent image inside the exposure time of $B_i$ and $M$ is the total number of reconstruction. Previous attempts reduce the discretization error by assuming linear \cite{chen2018reblur2deblur} or piece-wise linear motion \cite{red_xu2021motion} between latent frames and interpolating more intermediate frames, while this assumption might be violated in real-world scenarios especially in the case of complex non-linear motions. In contrast, we restore $\bar{L}_i[m]$ all by our network to exploit the real motion embedded in the event streams, and formulate the blurry-sharp loss $\mathcal{L}_{B\text{-}S}$ between the reblurred images $\bar{B}_i, \bar{B}_{i+1}$ and the original blurry inputs $B_{i}, B_{i+1}$ as
\begin{equation}\label{Blur-sharp}
    \mathcal{L}_{B\text{-}S} = \|\bar{B}_i - B_i \|_1 + \|\bar{B}_{i+1} - B_{i+1} \|_1,
\end{equation}
which guarantees the brightness consistency by learning from the blurry inputs.

\par 
\noindent \textbf{Sharp-event Loss.} 
Apart from the above constraints, the relation between sharp latent images and events can be also leveraged to supervise the reconstruction of consecutive latent frames. Based on Eq.~\eqref{latent_expEvent}, we have 
\begin{equation}
    \mathcal{N}(\Delta\tilde{L}) = \mathcal{N}(J),
\end{equation}
where 
$\Delta\tilde{L}\triangleq\tilde{L}(t) - \tilde{L}(f)$, $J\triangleq\int_f^t e(s) ds$ and $\mathcal{N}(\cdot)$ is the min/max normalization operator adopted in \cite{basics_paredes2021back}. Therefore, we can avoid the estimation of threshold $c$ and formulate the sharp-event loss $\mathcal{L}_{S\text{-}E}$ as
\begin{equation}
    \mathcal{L}_{S\text{-}E} = \|\mathcal{M}(\mathcal{N}(\Delta \tilde{L})) -  \mathcal{M}(\mathcal{N}(J)) \|_1,
\end{equation}
where $\mathcal{M}(\cdot)$ denotes a pixel-wise masking operator for $\mathcal{M}(\cdot) = 0$ only when there are no events. Finally, the total self-supervised framework can be summarized as follows.
\begin{equation}
  \mathcal{L} = \alpha \mathcal{L}_{B\text{-}E} + \beta \mathcal{L}_{B\text{-}S} + \gamma \mathcal{L}_{S\text{-}E},
\end{equation}
with $\alpha, \beta,\gamma$ denoting the balancing parameters.

\input{Figs-Fig-interp}
\begin{table*}[th]
\centering
\small
\caption{Quantitative results on the interpolation task. We compute PSNR and SSIM on the reconstruction results of the skipped frames, and use the official models provided by the authors for comparison. The column \#Param. indicates the number of network parameters.}
\vspace{-3mm}
\begin{tabular}{lcccccccccccc}
\hline
\multirow{3}{*}{Method} & \multicolumn{5}{c}{1 frame skip}                                &  & \multicolumn{5}{c}{3 frame skips}                              & \multirow{3}{*}{\#Param.} \\ \cline{2-6} \cline{8-12}
                        & \multicolumn{2}{c}{GoPro} &  & \multicolumn{2}{c}{HQF}          &  & \multicolumn{2}{c}{GoPro} &  & \multicolumn{2}{c}{HQF}         &                           \\ \cline{2-3} \cline{5-6} \cline{8-9} \cline{11-12}
                        & PSNR $\uparrow$       & SSIM $\uparrow$       &  & PSNR $\uparrow$          & SSIM $\uparrow$           &  & PSNR $\uparrow$       & SSIM $\uparrow$       &  & PSNR $\uparrow$          & SSIM $\uparrow$          &                           \\ \hline
Jin \cite{flawless_jin2019learning}                    & 20.47       & 0.5244      &  & 20.48          & 0.5958          &  & 19.50       & 0.4730      &  & 18.78          & 0.5160         & 10.81M                    \\
BIN \cite{bin_shen2020blurry}                    & 19.54       & 0.4645      &  & 18.25          & 0.4576          &  & -           & -           &  & -              & -              & 11.44M                    \\
DAIN \cite{dain_bao2019depth}                   & 20.89       & 0.5297      &  & 20.97          & 0.5980          &  & 20.48       & 0.5102      &  & 20.46          & 0.5848         & 24.03M                    \\
EDI \cite{edi_pan2019bringing}                   & 18.72       & 0.5059      &  & 16.62          & 0.4266          &  & 18.49       & 0.4862      &  & 16.58          & 0.4219         & -                         \\
LEDVDI \cite{ledvdi_lin2020learning}                 & \underline{24.42}       & \underline{0.8198}      &  & 19.24          & 0.6034          &  & \underline{23.57}       & \underline{0.7992}      &  & 18.57          & 0.5651         & \underline{4.996M}                    \\
Time Lens \cite{timelens_tulyakov2021time}               & 21.56       & 0.5809      &  & \underline{21.21}          & \underline{0.6090}          &  & 21.47       & 0.5870       &  & \underline{20.96}          & \underline{0.6060}         & 79.20M                    \\ \hline
\myname\ (Ours)               & \textbf{29.17}            & \textbf{0.8797}            &  & \textbf{23.09} & \textbf{0.6929} &  & \textbf{28.77}            & \textbf{0.8731}            &  & \textbf{22.24} & \textbf{0.6670} & \textbf{0.393M}           \\ \hline
\end{tabular}
\label{tab:interpolation}
\vspace{-3mm}
\end{table*}
\section{Experiments and Analysis}
\subsection{Experimental Settings}
\noindent \textbf{Datasets.} 
We evaluate the proposed method with three different datasets, including synthetic and real-world ones.



\par 
\textbf{GoPro:} We build a purely synthetic dataset based on the REDS dataset \cite{gopro_nah2019ntire}. We first downsample and crop the images to size $160\times320$ and then increase the frame rate by interpolating 7 images between consecutive frames using RIFE \cite{rife_huang2020rife}. Finally, we generate both blurry frames and events based on the high frame-rate sequences, where the blurry frames are obtained by averaging a specific number of sharp images, and events are simulated by ESIM \cite{esim_rebecq2018esim}.

\par 
\textbf{HQF:} The HQF dataset \cite{hqf_stoffregen2020reducing} contains real-world events and high-quality frames captured simultaneously by a DAVIS240C camera where the images are minimally blurred. We up-convert the frame rate and synthesize blurry frames using the same manner as the GoPro dataset, and form a semi-synthetic dataset of blurry videos.

\par 
\textbf{RBE:} The RBE dataset \cite{red_xu2021motion} employs a DAVIS346 camera to collect real-world blurry videos and the corresponding event streams, which can be used for training with the proposed self-supervised learning framework and verifying the performance of our method in real-world scenarios.


\par
\noindent \textbf{Implementation details.}
We implement the LDI network with 5 convolution layers and the fusion network with 6 convolution layers, 2 residual blocks and 1 CBAM \cite{cbam_woo2018cbam} block, forming a lightweight network architecture (detailed in the supplementary material). 
Our network is implemented using Pytorch and trained on NVIDIA GeForce RTX 2080 Ti GPUs with batch size 4 by default. The Adam optimizer \cite{kingma2014adam} is employed accompanied with the SGDR \cite{loshchilov2016sgdr} schedule where the parameter $T_{max}$ is set to 100 (reset the learning rate every 100 epochs). We set the number of temporal bins $N=16$ for LDI inputs and randomly crop the images to $128\times 128$ patches for training.
The training process is divided into two stages: 
We first train our model in the deblurring setting with the weighting factors $[\alpha,\beta,\gamma]=[512,1,1\times10^{-1}]$ and the initial learning rate $1\times10^{-3}$ for 100 epochs, and then continue training 
under the setting of unified deblurring and interpolation with the weighting factors $[\alpha,\beta,\gamma]=[128,1,1\times10^{-1}]$ and the initial learning rate $1\times10^{-4}$ for another 100 epochs. We train a model for each dataset and evaluate it on the corresponding dataset, which is convenient as we do not need ground-truth images for supervision. 

\subsection{Results of Deblurring}
For the setting of deblurring, we synthesize 1 blurry image using 49 frames on the GoPro and HQF datasets and evaluate the performance by restoring 7 original frames (before up-converting the frame rate) per blurry image. We compare to the state-of-the-art frame-based deblurring approach LEVS \cite{LEVS_jin2018learning} and event-based methods including EDI \cite{edi_pan2019bringing}, LEDVDI \cite{ledvdi_lin2020learning}, eSL-Net \cite{esl_wang2020event}, RED \cite{red_xu2021motion}, and evaluate the results by metrics PSNR, SSIM \cite{wang2003multiscale} and LPIPS \cite{zhang2018perceptual}.

\def\imgWidth{0.19\linewidth} 
\def\scc{(-0.6,-0.55)} 
\begin{figure}[!t]
\centering
\begin{overpic}[width=\imgWidth]{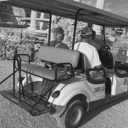}
\put(4,4){\footnotesize \textcolor{white}{\bf GT}}
\end{overpic}
\begin{overpic}[width=\imgWidth]{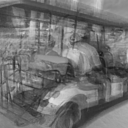}
\put(4,4){\footnotesize \textcolor{white}{\bf w/ $\mathcal{L}_{B\text{-}S}$}}
\end{overpic}
\begin{overpic}[width=\imgWidth]{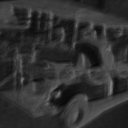}
\put(4,4){\footnotesize \textcolor{white}{\bf w/ $\mathcal{L}_{B\text{-}E}$}}
\end{overpic}
\begin{overpic}[width=\imgWidth]{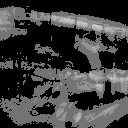}
\put(4,4){\footnotesize \textcolor{white}{\bf w/ $\mathcal{L}_{S\text{-}E}$}}
\end{overpic}
\begin{overpic}[width=\imgWidth]{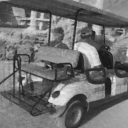}
\put(4,4){\footnotesize \textcolor{white}{\bf w/ All}}
\end{overpic}
\vspace{-0.5em}
\caption{Visual results of EVDI trained with different losses.}
\vspace{-1.5em}
\label{fig:ablation}
\end{figure}

\par 
As demonstrated in Tab.~\ref{tab:deblur}, the proposed method achieves remarkable deblurring results compared to the state-of-the-arts. 
The performance of LEVS is limited under highly dynamic scenes, \eg, Fig.~\ref{Fig-deblur}, due to the motion ambiguity.
For event-based approaches, the model-driven method EDI provides comparable performance to LEVS by exploiting the precise motion embedded in events. LEDVDI further enhances this advantage on the GoPro dataset through introducing learning-based techniques, but this overwhelming performance is not maintained in the HQF dataset due to the inconsistency between datasets.
RED achieves the most competing results with semi-supervised learning, but it still pays performance losses to balance different data distributions.
Our \myname\ method tackles this problem by fitting the particular data distribution with the self-supervised framework, and thus achieves the best performance on each dataset. 
Meanwhile, our model only contains 0.393M network parameters, which is an order of magnitude smaller than other methods except eSL-Net. Note that eSL-Net requires 122.8G FLOPs to infer a $160\times320$ image due to its recursive structure, while our model only needs 13.45G FLOPs, maintaining the overall efficiency.

\subsection{Results of Interpolation}
For the task of interpolation, we collect consecutive 97 frames (which are 13 original frames before up-converting) as a set of input, and synthesize 1 blurry images using 41 frames at both ends, leaving 1 latent original frame in the middle for evaluation (noted as 1 frame skip). Similarly, we design another case of 3 frame skips by synthesizing 1 blurry image with 33 frames and leaving 3 original middle frames. The frame-based interpolation methods Jin's work \cite{flawless_jin2019learning}, BIN \cite{bin_shen2020blurry}, DAIN \cite{dain_bao2019depth} and event-based approaches EDI \cite{edi_pan2019bringing}, LEDVDI \cite{ledvdi_lin2020learning}, Time Lens \cite{timelens_tulyakov2021time} are compared.

\par 
The results in Fig.~\ref{fig-interp} and Tab.~\ref{tab:interpolation} demonstrate the difficulty of blurry video interpolation for frame-based approaches. 
The optical flow used in DAIN often projects motion blur to the interpolation results as shown in Fig.~\ref{fig-interp}.
Jin's work employs a cascaded scheme for deblurring and interpolation, which tends to propagate the deblurring error to the interpolation stage. 
Despite BIN achieves joint deblurring and interpolation, it is difficult to synthesize the accurate intermediate frames due to the missing information between frames.
For event-based methods, LEDVDI and Time Lens are able to correctly estimate the intermediate frames using the precise motion inside events. However, the performance of Time Lens is highly dependent on the quality of reference frames, and LEDVDI often faces performance drop when inferring on other datasets due to data inconsistency. 
The proposed \myname\ method utilizes both frames and events to guarantee the interpolation quality, and tackles the inconsistency problem by learning on the target scenarios with the self-supervised framework, thus producing better results.

\begin{table}[!t]
\centering
\small
\caption{Ablation study of the proposed self-supervised framework and the fusion network. We train these models under the setting of 1 frame skip on the GoPro dataset but evaluate them by computing metrics on all frames of the test set, including the reconstruction results within and between blurry frames, for a comprehensive analysis of unified deblurring and interpolation.}
\vspace{-3mm}
\begin{tabular}{cccc|c}
\hline
$\mathcal{L}_{B\text{-}S}$ & $\mathcal{L}_{B\text{-}E}$ & $\mathcal{L}_{S\text{-}E}$ & Fusion & PSNR / SSIM / LPIPS \\ \hline
$\checkmark$      &       &       & $\checkmark$       & 22.66 / 0.6769 / 0.0954     \\
      & $\checkmark$      &       & $\checkmark$       & 9.152 / -0.0631 / 0.3213     \\
      &       & $\checkmark$      & $\checkmark$       & 6.847 / 0.0192 / 0.7840     \\
$\checkmark$      & $\checkmark$      &       & $\checkmark$       & \underline{29.97} / \underline{0.8998} / \underline{0.0182}     \\
$\checkmark$      &       & $\checkmark$      & $\checkmark$       & 28.07 / 0.8734 / 0.0274     \\
$\checkmark$      & $\checkmark$      & $\checkmark$      &        & 29.36 / 0.8924 / 0.0221     \\
$\checkmark$      & $\checkmark$      & $\checkmark$      & $\checkmark$      & \textbf{30.15} / \textbf{0.9026} / \textbf{0.0162}     \\ \hline
\end{tabular}
\label{tab:ablation}
\vspace{-5mm}
\end{table}
\subsection{Ablation Study}
We study the importance of each loss in our self-supervised framework and investigate the contribution of the fusion network. The following conclusions are drawn:
\par
\textbf{Necessity of loss combination.} 
As depicted in Fig.~\ref{fig:ablation}, 
blurry-sharp loss $\mathcal{L}_{B\text{-}S}$ contributes to brightness consistency but cannot produce sharp results. Blurry-event loss $\mathcal{L}_{B\text{-}E}$ and sharp-event loss $\mathcal{L}_{S\text{-}E}$ are able to deal with motion ambiguity by gaining supervision from blurry frames and events, respectively, but do not constrain brightness.
With the combination of loss functions, the brightness inconsistency and motion ambiguity can be simultaneously addressed by taking the complementary advantage of $\mathcal{L}_{B\text{-}S}$ and $\mathcal{L}_{B\text{-}E},\mathcal{L}_{S\text{-}E}$.


\par
\textbf{Importance of information fusion.} Although $\mathcal{L}_{B\text{-}E}$ and $\mathcal{L}_{S\text{-}E}$ are both capable of tackling motion ambiguity, their supervision comes from different information sources: $\mathcal{L}_{B\text{-}E}$ exploits blurry frames $B_i,B_{i+1}$ to supervise the estimation of $E(f,\mathcal{T}_i),E(f,\mathcal{T}_{i+1})$, while $\mathcal{L}_{S\text{-}E}$ utilizes events to constrain the generation of $\bar{L}(f)$. 
Hence, combing $\mathcal{L}_{B\text{-}E}$ and $\mathcal{L}_{S\text{-}E}$ will achieve the best performance, as demonstrated in Tab.~\ref{tab:ablation}.
Moreover, the fusion network also improves the results by fusing the information from different blurry frames and events.

\section{Conclusion}
This paper introduces a unified framework of event-based video deblurring and interpolation that generates high frame-rate sharp videos from low-frame-rate blurry inputs. Through analyzing the mutual constraints among blurry frames, sharp latent images and events, a self-supervised learning framework is also proposed to enable network training in real-world scenarios without any labeled data. Evaluation on both synthetic and real-world datasets demonstrates that our method competes favorably against state-of-the-arts while maintaining an efficient network design, showing potential for practical applications.





{\small
\bibliographystyle{ieee_fullname}
\bibliography{cvpr}
}

\end{document}